\documentclass[10pt,twocolumn,letterpaper]{article}

\usepackage{cvpr}
\usepackage{times}
\usepackage{epsfig}
\usepackage{graphicx}
\usepackage{amsmath}
\usepackage{amssymb}
\usepackage{multirow}
\usepackage{algorithm}
\usepackage{algpseudocode}

\usepackage{float}

\usepackage[breaklinks=true,bookmarks=false]{hyperref}

\usepackage{array}
\newcolumntype{L}[1]{>{\raggedright\let\newline\\\arraybackslash\hspace{0pt}}m{#1}}
\newcolumntype{C}[1]{>{\centering\let\newline\\\arraybackslash\hspace{0pt}}m{#1}}
\newcolumntype{R}[1]{>{\raggedleft\let\newline\\\arraybackslash\hspace{0pt}}m{#1}}

\cvprfinalcopy % *** Uncomment this line for the final submission

\newcommand{\Figure}[1]{fig.~\ref{#1}}
\newcommand{\Section}[1]{\S\ref{#1}}
\newcommand{\tableref}[1]{table~\ref{#1}}
\newcommand{\Tableref}[1]{Table~\ref{#1}}
\newcommand{\Algorithm}[1]{Algorithm~\ref{#1}}

\begin{document}

%%%%%%%%% TITLE
\title{Multiple Instance Reinforcement Learning for \\ Efficient Weakly-Supervised Detection in Images}

% Pages are numbered in submission mode, and unnumbered in camera-ready
\ifcvprfinal\pagestyle{empty}\fi

\newcommand*\samethanks[1][\value{footnote}]{\footnotemark[#1]}
\author{
	\ Stefan Mathe$^{1,3}$
	\hspace{0.1cm} and \hspace{0.1cm}
	%\hspace{1.0cm}
	Cristian Sminchisescu$^{2,1}$
	\\
$^1$Institute of Mathematics of the Romanian Academy of Science\\
$^2$Department of Mathematics,  Faculty of Engineering, Lund University \\
$^3$Department of Computer Science, University of Toronto \\
	{\tt\small stefan.mathe@imar.ro,  cristian.sminchisescu@math.lth.se}
	\\
}

\vspace{-0.3cm}

%\author{Stefan Mathe\\
%Institute of Mathematics at the Romanian Academy\\
%Institution1 address\\
%{\tt\small firstauthor@i1.org}
%\and
%Second Author\\
%Institution2\\
%First line of institution2 address\\
%{\tt\small secondauthor@i2.org}
%}

\def\balpha{\boldsymbol{\alpha}}
\def\bdelta{\boldsymbol{\delta}}
\def\btheta{\boldsymbol{\theta}}
\def\bxi{\boldsymbol{\xi}}

\vspace{-1cm}

\maketitle

\vspace{-1cm}

%%%%%%%%% ABSTRACT
\begin{abstract}
State-of-the-art visual recognition and detection systems increasingly rely on large amounts of training data and complex classifiers. Therefore it becomes increasingly expensive both to manually annotate datasets and to keep running times at levels acceptable for practical applications. In this paper, we propose two solutions to address these issues. First, we introduce a weakly supervised, segmentation-based approach to learn accurate detectors and image classifiers from weak supervisory signals that provide only approximate constraints on target localization. We illustrate our system on the problem of action detection in static images (Pascal VOC Actions 2012), using human visual search patterns as our training signal. Second, inspired from the saccade-and-fixate operating principle of the human visual system, we use reinforcement learning techniques to train efficient search models for detection. Our sequential method is \emph{weakly supervised and general} (it does not require eye movements), finds optimal search strategies for any given detection confidence function and achieves performance similar to exhaustive sliding window search at a fraction of its computational cost.
\end{abstract}

%%%%%%%%% BODY TEXT

\section{Introduction}

The central problem of constructing an object detector can be decomposed into learning a confidence detection response function and estimating a search strategy. Most frequently, confidence functions are learned in a fully supervised setting, and the search is performed exhaustively. This approach is however not well suited to contemporary state-of-the-art systems, which are trained using large amounts of image or video data, and require complex multi-layer models\cite{KrizhevskyEtAl-nips-2012}. First, manually annotating the large amounts of data needed by supervised algorithms is increasingly expensive. This underlines the need to exploit alternative sources of information to support system accuracy and efficiency. Second, the target search complexity is sometimes high in many practical object detection applications.

Human eye movements can provide a rich source of supervision which, due to recent developments in eyetracking hardware, is increasingly less invasive and expensive. Large scale datasets annotated with human eye movements under various task constraints have recently been acquired and made public\cite{MatheSminchisescu-eccv-2012,MatheSminchisescu-nips-2013,PapadopoulosEtAl-eccv-2014}. Such data has been exploited at training time to support visual classification performance in video\cite{MatheSminchisescu-eccv-2012}, but its usefulness as a training signal for detection, in the absence of any additional annotations, remains unexplored. Nor has detection performance benefited from insights derived from the `saccade-and-fixate' operating principles of the human visual system.

In this paper, we propose general methods to learn detector confidence functions and search models from weak supervisory signals, circumventing the need for manual image annotations. Our main contributions are:

\begin{itemize}
\item We demonstrate a novel segmentation-based constrained multiple instance model to learn effective action detectors in static images. Our method is trained using image labels and associated eye movement data and requires no ground truth annotations of the spatial extent of the targets, \textit{e.g.} bounding boxes or image segments (see \Section{s:confidence_learning} and tables \ref{t:detap},\ref{t:detap_inclusion},\ref{t:classification}). 
\item We show, for the first time, that human eye movement information can be integrated with image labels to train more accurate image classification pipelines. The performance of our system approaches that of pipelines trained under the stronger supervision of target bounding box information (see \Section{s:confidence_learning} and \tableref{t:classification}).
\item We develop a weakly supervised reinforcement learning methodology to obtain optimal sequential models that operate in a fixate-and-saccade regime. Our method  achieves significant improvements in search efficiency at virtually the same detection and classification accuracy as exhaustively evaluated sliding window methods (see \Section{s:sequential_learning} and tables \ref{t:detap},\ref{t:detap_inclusion},\ref{t:classification},\ref{t:seq_det_speedup}).
\end{itemize}

\section{Related Work}

Many methods have been proposed to accelerate detectors. Prominent techniques are based on branch-and-bound heuristics\cite{Kokkinos-iccv-2013,LampertEtAl-cvpr-2008}, hierarchies of classifiers\cite{VedaldiEtAl-iccv-2009} or methods that reuse computation between neighboring regions\cite{WeiTao-cvpr-2010}. In turn, different features have been used in the design of the detector response functions. Deep convolutional neural networks have surpassed methods based on support vector machines on many computer vision problems, such as the image classification\cite{KrizhevskyEtAl-nips-2012}, object classification\cite{OquabEtAl-cvpr-2014}, object detection\cite{GirschickEtAl-cvpr-2014}, action classification\cite{OquabEtAl-cvpr-2014} and pose prediction\cite{ToshevEtAl-cvpr-2014}. Multiple instance learning\cite{DietterichEtAl-ai-1997} formulations can be seen as a generalization of supervised learning, in which class labels are assigned to sets of training examples. Many algorithmic solutions have been proposed, based on SVMs\cite{AndrewsEtAl-nips-2002,BunescuMooney-icml-2007}, CRFs\cite{VezhnevetsEtAl-cvpr-2012} or boosted classifiers\cite{AliSaenko-cvpr-2014} (see \cite{Amores-ai-2013} for a review).

Datasets of human eye movement have been collected for both images and video. Search targets include pedestrians\cite{EhingerEtAl-viscog-2009}, faces \cite{CerfEtAl-nips-2007} and actions\cite{MatheSminchisescu-eccv-2012,MatheSminchisescu-nips-2013,PapadopoulosEtAl-eccv-2014}. 
%Human visual search under natural stimuli has been studied both in terms of visual saliency\cite{EhingerEtAl-viscog-2009,MatheSminchisescu-eccv-2012,CerfEtAl-nips-2007} and sequential ordering\cite{MatheSminchisescu-nips-2013}. 
Eye movements have successfully been used to boost the performance of computer vision systems, such as action classification from video\cite{MatheSminchisescu-eccv-2012,VigEtAl-eccv-2012,FathiEtAl-eccv-2012}, action detection from video\cite{ShapovalovaEtAl-nips-2013} or image segmentation\cite{RamanathanEtAl-eccv-2010}. Some of these systems assume the availability of eye movement data at test time\cite{VigEtAl-eccv-2012,RamanathanEtAl-eccv-2010}, while some do not\cite{MatheSminchisescu-eccv-2012,FathiEtAl-eccv-2012,ShapovalovaEtAl-nips-2013}. Human eye movements have also been recently used to learn object detectors by Papadopoulos \textit{et al.}\cite{PapadopoulosEtAl-eccv-2014}. Their method employs supervised learning techniques to predict a bounding box given the human's scanpath, then reduces to a classical supervised detection learning problem, in which the predicted bounding boxes play the role of ground truth annotations. However, ground truth bounding box information must still be available, at least for a subset of the training images. This can be a  limiting factor in many practical applications. In this work, we take a markedly different approach. Instead of splitting the problem into two learning steps, we treat it as a joint optimization problem, recovering both the ground truth regions and the confidence function in a single stage. Unlike \cite{PapadopoulosEtAl-eccv-2014}, our method does not require the availability of ground truth bounding boxes at training time.

Visual analysis systems based on fixate-and-saccade ideas have been proposed in \cite{LarochelleHinton-nips-2009,ButkoMovellan-tamd-2010,MatheSminchisescu-nips-2013}, for the problem of digit recognition\cite{LarochelleHinton-nips-2009}, face detection\cite{ButkoMovellan-tamd-2010} and saccade prediction\cite{MatheSminchisescu-nips-2013}. Here, we derive novel, fully trainable models for two challenging problems, namely action recognition and detection in cluttered natural scenes, and in a reinforcement optimal learning setup\cite{SuttonBarto-1998}.

\begin{figure}
\begin{center}
\scalebox{0.36}{
\includegraphics[viewport=4.0cm 30.5cm 24.3cm 41cm]{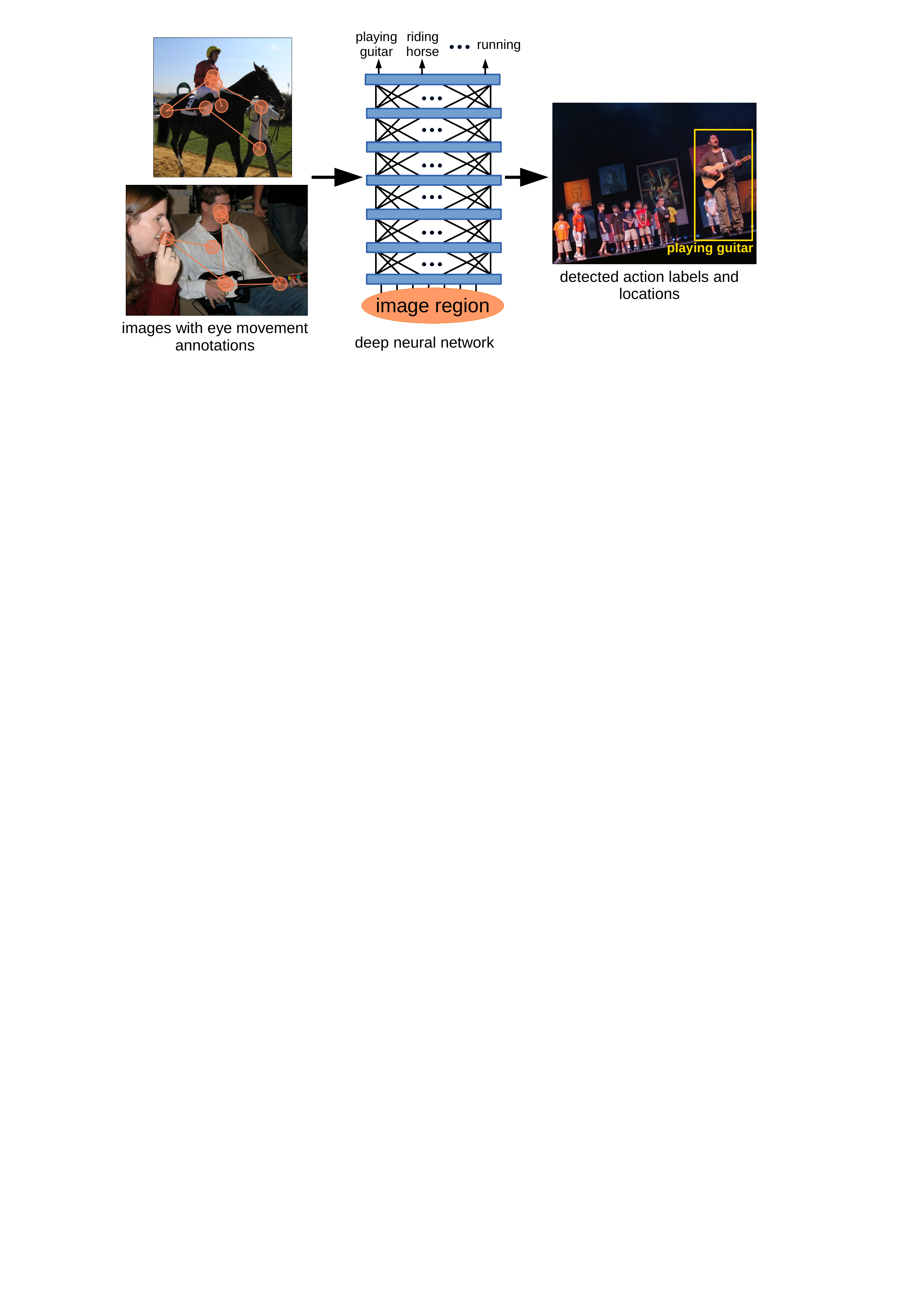}
}
\end{center}
\caption{We propose to learn action detectors from images annotated with human eye movements. Note that no bounding boxes or image segmentation annotations are required at training time.}\label{fig:system_outline}
\end{figure}

\section{Action Detection in Static Images}

Given an image showing one or more persons, we want to determine both the actions present (from a predefined list) and their corresponding spatial support. The spatial support is more difficult to objectively define for an action, than it is for an object. Here, we consider the spatial support of an action to be a tight bounding box that encloses both the human actor and the additional objects the action requires. For example, the spatial support for an instance of the action \textit{playing guitar} is a bounding box that tightly encloses both the artist and the instrument (see \Figure{fig:system_outline}).

We are not aware of any publicly available dataset suited for this problem. While the Pascal VOC Action Classification Challenge\cite{pascal-voc-2012} contains a large set of images with human actions, the provided annotations are not useful for action detection for several reasons. First, not all action instances are annotated. Second, many images contain crowds of people performing actions (\textit{e.g.} images from running competitions), with many overlapping targets. These images are incompletely annotated and make the evaluation of an action detector subjective. Third, not all actions are suitable for detection given the image statistics: the extent of many actions is often the entire image, making spatial localization uninteresting (\textit{e.g.} \textit{playing piano} or \textit{using desktop computer}).

We address this problem by providing our own complete annotations for a subset of the PASCAL VOC Actions 2012 database\footnote{Our annotations will be made publicly available upon publication.}. Our dataset contains 4412 images and 10 action classes: \textit{jumping}, \textit{phoning}, \textit{playing guitar}, \textit{reading}, \textit{riding bike}, \textit{riding horse}, \textit{running}, \textit{walking} and \textit{using laptop}. Note that our action classes are slightly different than in the PASCAL VOC Actions dataset. We have specialized the classes \textit{playing instrument} and \textit{using computer}, to keep the extent of the action within bounds that make detection non-trivial. We have also removed all images containing crowds that perform the same action, but are only sparsely annotated.

\section{Problem Formulation}

Given an input image, we formulate action detection as the problem of maximizing a confidence function $f_{\text{c}}:R\rightarrow\mathbb{R}$ over the set of image regions $R$:
\begin{equation}\label{eq:detection}
\mathbf{r}^{*}=\arg\max_{\mathbf{r}\in R} f_{\text{c}}(\mathbf{r})
\end{equation}

The set of image regions $R$ can be defined either at the coarse level of bounding boxes or at the finer level of image segments. In the present work, the region space $R$ consists of all segments extracted using the CPMC algorithm\cite{CarreiraSminchisescu-pami-2012}. There are two reasons that motivate our choice. First, the space of image segments is several orders of magnitude smaller than the one of bounding boxes. Second, figure-ground image segments produced by CPMC can be mapped with reasonable accuracy to objects or groups of related objects that represent human actions (see \tableref{t:segment_filtering}, column 3). Furthermore, as shown in \Section{s:fixation_studies}, humans often fixate segments that well match the target during visual search. Hence, for action detection, image segments provide a good tradeoff between computational complexity and spatial descriptive power.

Typically, the confidence function $f_{\text{c}}$ is learnt from a training set of images annotated with ground truth regions containing the target instances (usually bounding boxes). This type of annotation is however expensive to obtain in practice. An alternative, cheaper `annotation' consists of sequences of eye movements made by human subjects searching for the target, together with a label indicating the presence or absence of the target class in images. In \Section{s:confidence_learning}, we present a method to learn the confidence function $f_c$ from this information only, without the need for region-level annotations.

When looking at an image, the human visual system exhibits efficient search strategies that only explore a few locations before deciding on the presence of a target. When asked to locate an action, humans take on average only $3.6$ fixations (stdev=$1.8$) in the PASCAL VOC Actions 2012 dataset. In \Section{s:sequential_learning} we present a model to learn efficient search strategies in an weakly supervised setting, optimally formulated in a reinforcement learning setup. Our model operates in a fixate-and-saccade regime, inspired from the workings of the human visual system.

\section{Learning Confidence Functions using Eye Movement Annotations}\label{s:confidence_learning}

In the classical (fully supervised) approach, the confidence function $f_{\text{c}}$ is the response of a classifier, trained to separate ground truth image regions from a set of regions having little or no overlap with the positive region set. In this section we consider the more challenging setup in which no ground truth image regions are available at training time. Instead, our training signal consists of image labels -- indicating the presence or absence of an action in the image -- and eye movements recorded from human subjects asked to locate instances of the target action (see \Figure{fig:system_outline}). Such data has been recently acquired and made publicly available\cite{MatheSminchisescu-nips-2013}.

In \Section{s:fixation_studies} we begin with a brief analysis of the localization power of human eye movements. In \Section{s:confidence_learning_method}, we describe our method in detail. We evaluate the model and discuss experimental results in \Section{s:confidence_learning_results}.

\subsection{Human Fixations and Action Localization}\label{s:fixation_studies}

The sequence of fixations and saccades made by a human subject is precisely registered with the image but contains weaker information on the location and extent of the search target compared to a bounding box or a segment. Consider for example the case of a human presented with an image of a concert scene where one of the artists is playing a guitar. Typically the last few fixations fall onto relevant parts of the target, \textit{i.e.} the person and the musical instrument. However, the remaining part of the scanpath is most likely focused on the background and other objects or humans.

\begin{table}
\scalebox{0.78}{
\begin{tabular}{|c|c|c|c|}
\hline
\multirow{3}{*}{\textbf{image label}}
&
\multirow{3}{2.0cm}{\centering\textbf{percent of fixated segments}}
&
\multicolumn{2}{c|}{\multirow{2}{5.0cm}{\centering\textbf{best overlap with ground truth}}}
%&
%\multirow{3}{2.2cm}{\centering\textbf{best overlap with ground truth}} 
\\
 & & \multicolumn{2}{c|}{} \\
% & & \multicolumn{2}{c|}{} \\
\cline{3-4}
 & & \textbf{fixated segments} & \textbf{all segments} \\
\hline
\textbf{jumping}        & $36.7\%$ & $70.0\%$ & $70.7\%$ \\
\hline
\textbf{phoning}        & $33.1\%$ & $79.2\%$ & $81.4\%$ \\
\hline
\textbf{playing guitar} & $32.8\%$ & $82.7\%$ & $83.9\%$ \\
\hline
\textbf{reading}        & $34.7\%$ & $74.7\%$ & $77.3\%$ \\
\hline
\textbf{riding bike}    & $32.3\%$ & $73.9\%$ & $74.7\%$ \\
\hline
\textbf{riding horse}   & $33.5\%$ & $78.2\%$ & $78.7\%$ \\
\hline
\textbf{running}        & $37.0\%$ & $66.6\%$ & $67.3\%$ \\
\hline
\textbf{taking photo}   & $33.2\%$ & $76.1\%$ & $78.2\%$ \\
\hline
\textbf{walking}        & $34.6\%$ & $56.9\%$ & $58.2\%$ \\
\hline
\textbf{using laptop}   & $32.9\%$ & $79.7\%$ & $80.8\%$ \\
\hline
\hline
\textbf{mean}           & $\mathbf{34.1}\%$ & $\mathbf{73.8}\%$ & $\mathbf{75.1}\%$\\
\hline
\end{tabular}
}
\vspace{-1mm}
\caption{When searching for an action, humans fixate approximately one third of the segments in an image (column 1).
At the same time, the set of fixated segments contains elements which overlap the search taget well (column 2), close to the best attainable overlap given the entire segment pool (column 3). This suggests that fixations represent a strong filtering signal over this pool.}\label{t:segment_filtering}
\vspace{-2mm}
\end{table}

How weak is the localization signal provided by human fixations? \Tableref{t:segment_filtering} lists the percentage of fixated image segments in the Pascal VOC Actions 2012 dataset. It also shows the maximum overlap between a fixated image segment and the ground truth bounding box. Both figures are averaged over all images belonging to an action class. The  analysis suggests that, while fixations do not give precise information on the location of the search target, they can serve as a strong filtering signal over the entire segment pool.

\subsection{Method description}\label{s:confidence_learning_method}

Based on the analysis in \Section{s:fixation_studies}, we develop a multiple instance learning formulation, in which training instances are the image regions $\mathbf{r}_i$ obtained by a segmentation algorithm on the training set, referenced in a global index set $I$. In a one-vs-all formulation, training image with index $j \in J$ is represented by {\it(1)} a bag $B_j \subset I$ of region indexes belonging to image $j$ and {\it(2)} an associated image label $l_j \in \left\{-1,1\right\}$.

A positive bag $B_j$ corresponds to an image where the target action is present and contains all regions fixated by humans viewing that image. Some of these regions represent the detection target, and some do not. Negative bags $B_j$ correspond to images where the action is absent, and contain all regions in that image (fixated or not). We know that none of these regions represents the detection target.

We wish to simultaneously find the optimal assignments $y_i$ of labels to instances and the optimal separating hyperplane for the positive and negative instances under this assignment, subject to several constraints:

\noindent\textbf{Positivity constraints:} There should be at least one positive instance inside each positive bag, and no positive instances inside negative bags.

\noindent\textbf{Inclusion constraints:} Positivity constrains implicitly encourage good separation of positive bags from negative bags, under their maximum classifier response. This does not imply, however, good localization performance.

Consider, for example, a classifier that responds to the presence of some pattern strongly associated to the target, \textit{anywhere} inside the input region. This classifier cannot distinguish among image regions containing the target. It will separate positive and negative bags well, but will have poor localization capabilities. We deal with this issue by imposing that, for any region that is positively labeled, all its subregions $S(\textbf{r})$ be labeled negative. We define the set $S(\textbf{r})$ using a soft threshold criterion:
\vspace{-2mm}
\begin{equation}\label{eq:subseg}
S(\mathbf{r})=\left\{\mathbf{r'} \;\text{s.t.}\; \left|\mathbf{r} \cap \mathbf{r}' \right|/\left|\mathbf{r}'\right| \leq T_{S} \right\}
\end{equation}

\vspace{-3mm}

\noindent\textbf{Unconstrained fringe:} Previous experience with training sliding window detectors\cite{OquabEtAl-cvpr-2014} has shown that excluding bounding boxes overlapping the ground truth from the set of negative examples can greatly improve the localization power. We define the \textit{fringe} $\text{F}(\mathbf{r})$ of region $\textbf{r}$ as the set of regions that overlap $\textbf{r}$, down to a threshold:
\vspace{-2mm}
\begin{equation}\label{eq:fringe}
F(\mathbf{r})=\left\{\mathbf{r'} \;\text{s.t.}\; \left(\mathbf{r'} \neq \mathbf{r}\right) \wedge \left(\left|\mathbf{r} \cap \mathbf{r}' \right|/\left|\mathbf{r} \cup \mathbf{r}'\right| \geq T_{F}\right) \right\}
\end{equation}

\vspace{-3mm}

In our formulation, regions in the fringe of instances selected as positive do not affect the separating hyperplane and their corresponding labels are constrained to be $0$.

This leads to the following formulation:
\begin{equation}\label{eq:cmisvm}
\min_{y_i} \min_{\mathbf{w},b,\bxi} \frac{1}{2} \left\|\mathbf{w}\right\|^2 + C \sum_i \xi_i
\end{equation}

\vspace{-3mm}

\noindent s.t. \scalebox{0.90}{$\forall i \in I: \; y_i \in \left\{-1,0,1\right\} \hspace{4.9cm}(4.1)$}

\noindent \hspace{0.43cm} \scalebox{0.90}{$\forall i \in I: \; y_i=1 \implies \left(\forall k\right) \in F(\mathbf{r}_i), y_k=0 \hspace{2.2cm}(4.2)$ }

\noindent \hspace{0.43cm} \scalebox{0.90}{$\forall i \in I: \; y_i=1 \implies \left(\forall k\right) \in S(\mathbf{r}_i)\setminus F(\mathbf{r}_i), y_k=-1 \hspace{0.84cm}(4.3)$ }

\noindent \hspace{0.43cm} \scalebox{0.90}{$\forall i \in I: \; y_i\neq 0 \implies y_i\left(\mathbf{w}^T g\left(\mathbf{r}_i\right)+b\right) \geq 1 - \xi_i,\;\xi_i>0 \hspace{0.23cm}(4.4)$}

\noindent \hspace{0.43cm} \scalebox{0.90}{$\forall j \in J: \; l_j=1 \implies \left(\exists\right) i \in B_i \;\text{s.t.}\; y_i=1 \hspace{2.25cm}(4.5)$}  %\sum_{j \in B_i} \frac{y_j+1}{2}\geq 0$}

\noindent \hspace{0.43cm} \scalebox{0.90}{$\forall j \in J: \; l_j=-1 \implies(\forall) i \in B_j, y_i=-1 \hspace{2.2cm}(4.6)$}

\vspace{2mm}
\noindent where $g:R \rightarrow \mathbb{R}^{n}$ is a feature extractor that takes as input an image region.

Note that if we remove constraints (4.2), (4.3), and the value $0$ for the range of possible image labels from constraint (4.1), we end up with the miSVM formulation\cite{AndrewsEtAl-nips-2002}, which treats all training instances as independent from one another.

\begin{algorithm}
\caption{Constained Multiple Instance SVM learning}\label{alg:cmisvm}
\begin{algorithmic}[1]
\Procedure{CMI-SVM}{$\left\{\mathbf{r}_i\right\}_{i \in I},\left\{\mathbf{B}_j,l_j\right\}_{j \in J}$}
\State $y_i \leftarrow \text{rand}(\left\{-1,1\right\})$
\Repeat
	\State $(\mathbf{w},b) \leftarrow \text{SVM}(\left\{\mathbf{r}_i,y_i\right\}_{y_i\neq 0})$
	\For{$j \in J$}
		\If{$l_j=1$}
			\State $V \leftarrow B_j$;\; $n_{\text{pos}}\leftarrow0$
			\Repeat
				\State $r \leftarrow \max_{i \in V} \mathbf{w}^T g\left(\mathbf{r}_i\right)$
				\State $k \leftarrow \arg\max_{i \in V} \mathbf{w}^T g\left(\mathbf{r}_i\right) $
				\If{$r>0$}
					\State $y_k \leftarrow 1$;\;$n_{\text{pos}} \leftarrow n_{\text{pos}}+1$
					\State $y_i \leftarrow -1$, for each $i \in S(\mathbf{r}_k)$
					\State $y_i \leftarrow 0$, for each $i \in F(\mathbf{r}_k)$
					\State $V \leftarrow V \setminus \left(F(\mathbf{r}_k) \cup S(\mathbf{r}_k)\right)$
				\Else
					\State $y_k \leftarrow 0$
					\State $V \leftarrow V \setminus \left\{k\right\}$
				\EndIf
			\Until{$V=\emptyset$}
			\If{$n_{\text{pos}}=0$}
				\State $k \leftarrow \arg\max_{i \in B_j} \mathbf{w}^T g\left(\mathbf{r}_i\right) $
				\State $y_k \leftarrow 1$
			\EndIf
		\Else
			\State $y_i \leftarrow -1$, for each $i \in B_j$
		\EndIf
	\EndFor
\Until{$\mathbf{y}$ has not changed during the iteration}
\State \Return $\mathbf{w},b,\mathbf{y}$
\EndProcedure
\end{algorithmic}
\end{algorithm}

We solve the optimization problem \eqref{eq:cmisvm} using an iterative procedure, presented in \Algorithm{alg:cmisvm}. We alternate between finding the SVM parameters $\mathbf{w}$ and $b$ for the current assignment estimate $\mathbf{y}$ (line 4), and determining $\mathbf{y}$ given parameters, subject to imposed constraints (lines 5-28).

For each bag $j$ with positive label, we iteratively look for a yet unlabeled region with maximal positive SVM response (lines 9-10). If found, we label it $1$, its fringe $0$ and all its subregions $-1$ (lines 11-15). Otherwise, all remaining regions are labeled as negative (lines 16-19). If no positive SVM responses are generated inside a positive bag, we enforce consistency by labeling the region with maximal response as 1 (lines 21-24). Instances in negative bags are always labeled negative (lines 25-27). We detect convergence when bag labels no longer change between iterations.

\section{Weakly Supervised Reinforcement Learning of Search Strategies for Detection}\label{s:sequential_learning}

Equation \eqref{eq:detection} involves an exhaustive search over the entire set $R$ image regions in the test image. In this section, we propose a weakly supervised model that aims to minimize the computational load needed to locate the target, by restricting the search to a subset of $R$. We present our model in \Section{s:sequential_learning_model}. The methodology and experimental results are discussed in \Section{s:sequential_learning_results}.

\begin{algorithm}
\caption{Policy sampling algorithm.}\label{alg:policy_sampling}
\begin{algorithmic}[1]
\Procedure{sample }{$\mathbf{s}_t=\left(H_{t},S_{t}\right)$}
\State $c_t \leftarrow \max_{i \in H_t} f_{\text{c}}(\mathbf{r}_i)$
\State $d_t \sim  p(d_t|\mathbf{s}_t)$ using \eqref{eq:prob_action}

%\textbf{if}$d_t=1$\textbf{then}$k \leftarrow \arg\max_{i \in H_t} f_{\text{c}}(\mathbf{r}_i)$

\If{$d_t=1$}\State $k \leftarrow \arg\max_{i \in H_t} f_{\text{c}}(\mathbf{r}_i)$
  
  \State Done. Predict region $\mathbf{r}_k$ with confidence $c_t$.
  %\State \Return $\left(\text{done},c,\mathbf{r}^{*}\right)$
\Else
  \State $e_{t} \sim p(e_t|\mathbf{s}_t)$ using \eqref{eq:prob_evidence}
  \State $\mathbf{z}_{t} \sim p(\mathbf{z}_t|\mathbf{s}_t,e_{t})$ using \eqref{eq:prob_position}
  \State \Return $\mathbf{a}_t=\left(e_t,\mathbf{z}_t\right)$
\EndIf
\EndProcedure
\end{algorithmic}
\end{algorithm}

%\begin{figure*}
%\begin{center}
%\scalebox{0.4}{\includegraphics[viewport=4cm 13cm 12cm 25cm]{graphical_model.pdf}}
%\end{center}
%\caption{Sequential detector action sampling algorithm \textbf{(a)} and the underlying graphical model. \textbf{(b)}}
%\end{figure*}
\vspace{-4mm}
\begin{algorithm}
\caption{State transition algorithm.}\label{alg:transition}
\begin{algorithmic}[1]
\Procedure{observe }{$\mathbf{s}_t=\left(H_{t},S_{t}\right),\;\mathbf{a}_t=\left(e_t,\;\mathbf{z}_t\right)$}
\State $O_{t} \leftarrow \left\{i \in B \mid \mathbf{z}_t \in \mathbf{r}_i \right\}$
%\State $H_{t+1} \leftarrow H_t \cup O_t$
%\State $S_{t+1} \leftarrow \left(S_t \setminus \left\{e_t\right\}\right) \cup O_t$
\State \Return $\mathbf{s}_{t+1}=\left(H_t \cup O_t,S_t \cup \left\{e_t\right\}\right)$
%$p(\mathbf{s}_{t+1} | \mathbf{s}_t,\mathbf{a}_t)$
%$	r_t$
\EndProcedure
\end{algorithmic}
\end{algorithm}

\begin{table*}
\begin{center}
\scalebox{0.78}{
\begin{tabular}{|c|c|c|c|c||c|c|}
\hline
\multirow{2}{*}{\textbf{label}}     & \multicolumn{6}{c|}{\textbf{method/annotations}} \\
\cline{2-7}
%                        &  \textbf{image labels}       & \textbf{eye movements+image labels} & \textbf{labeled bounding boxes} & \textbf{labeled bounding boxes}  \\
                        &  \textbf{CMI-IL-DET}         & \textbf{MI-EYE-DET} & \textbf{CMI-EYE-DET}                & \textbf{BB-DET}                 & \textbf{CMI-EYE-SEQ}   &  \textbf{BB-SEQ} \\
\hline
\textbf{jumping}        &   $2.0\%$                    & $4.0\%$             & $2.5\%$                             & $38.4\%$                        & $3.5\% \pm 0.1\%$  & $35.6\% \pm 1.0\%$ \\
\hline
\textbf{phoning}        &   $0.7\%$                    & $13.5\%$            & $16.8\%$                            & $41.8\%$                        & $14.7\% \pm 0.2\%$ & $39.1\% \pm 0.9\%$ \\
\hline
\textbf{playing guitar} &   $7.1\%$                    & $19.6\%$            & $33.0\%$                            & $42.2\%$                        & $26.6\% \pm 1.1\%$ & $47.0\% \pm 0.1\%$ \\
\hline
\textbf{reading}        &   $0.6\%$                    & $8.0\%$             & $10.2\%$                            & $17.6\%$                        & $8.4\% \pm 0.2\%$  & $17.2\% \pm 0.2\%$ \\
\hline
\textbf{riding bike}    &   $3.8\%$                    & $8.1\%$             & $11.8\%$                            & $45.7\%$                        & $7.4\% \pm 0.1\%$  & $42.1\% \pm 0.4\%$ \\
\hline
\textbf{riding horse}   &   $3.3\%$                    & $25.9\%$            & $45.6\%$                            & $69.0\%$                        & $36.5\% \pm 0.5\%$ & $65.4\% \pm 0.5\%$ \\
\hline
\textbf{running}        &   $0.2\%$                    & $1.7\%$             & $1.6\%$                             & $36.7\%$                        & $1.4\% \pm 0.2\%$  & $27.6\% \pm 1.4\%$ \\
\hline
\textbf{taking photo}   &   $0.6\%$                    & $4.5\%$             & $7.8\%$                             & $15.1\%$                        & $7.2\% \pm 0.1\%$  & $15.4\% \pm 1.0\%$ \\
\hline
\textbf{walking}        &   $0.5\%$                    & $0.6\%$             & $1.3\%$                             & $23.6\%$                        & $0.2\% \pm 0.0\%$  & $15.5\% \pm 1.4\%$ \\
\hline
\textbf{using laptop}   &   $1.0\%$                    & $10.7\%$            & $14.6\%$                            & $39.2\%$                        & $9.2\% \pm 0.2\%$  & $39.3\% \pm 0.2\% $ \\
\hline
\hline
\textbf{mean}           &   $\mathbf{2.1}\%$           & $\mathbf{9.7\%}$     & $\mathbf{14.5\%}$                   & $\mathbf{36.9\%}$              & $\mathbf{11.5\% \pm 0.1\%}$ & $\mathbf{34.4\% \pm 0.3\%}$ \\
\hline
\end{tabular}
}
\end{center}
\vspace{-2mm}
\caption{Detection performance (VOC detection average precision criterion) for different methods and levels of supervision.}\label{t:detap}
\vspace{-3mm}
\end{table*}

\begin{table*}
\begin{center}
\scalebox{0.78}{
\begin{tabular}{|c|c|c|c|c||c|c|}
\hline
\multirow{2}{*}{\textbf{label}}     & \multicolumn{6}{c|}{\textbf{method/annotations}} \\
\cline{2-7}
%                        &  \textbf{image labels}       & \textbf{eye movements+image labels} & \textbf{labeled bounding boxes} & \textbf{labeled bounding boxes}  \\
                        &  \textbf{CMI-IL-DET}         & \textbf{MI-EYE-DET} & \textbf{CMI-EYE-DET}                & \textbf{BB-DET}                 & \textbf{CMI-EYE-SEQ} & \textbf{BB-SEQ} \\
\hline
\textbf{jumping}        &   $18.1\%$                   & $12.1\%$    & $9.1\%$                            & $44.1\%$                                 & $9.6\% \pm 0.3\%$  & $39.7\% \pm 1.3\%$ \\
\hline
\textbf{phoning}        &   $20.6\%$                   & $34.3\%$    & $41.2\%$                           & $48.6\%$                                 & $38.3\% \pm 0.6\%$ & $50.3\% \pm 0.8\%$ \\
\hline
\textbf{playing guitar} &   $36.8\%$                   & $47.3\%$    & $57.5\%$                           & $52.2\%$                                 & $52.5\% \pm 0.9\%$ & $49.7\% \pm 0.1\%$ \\
\hline
\textbf{reading}        &   $11.8\%$                   & $15.9\%$    & $27.7\%$                           & $25.0\%$                                 & $25.7\% \pm 0.5\%$ & $24.3\% \pm 0.4\%$ \\
\hline
\textbf{riding bike}    &   $32.0\%$                   & $21.1\%$    & $27.8\%$                           & $55.3\%$                                 & $27.4\% \pm 0.1\%$ & $57.3\% \pm 1.6\%$ \\
\hline
\textbf{riding horse}   &   $35.6\%$                   & $37.9\%$    & $52.7\%$                           & $78.7\%$                                 & $54.9\% \pm 0.8\%$ & $73.3\% \pm 1.0\%$ \\
\hline
\textbf{running}        &   $13.9\%$                   & $8.2\%$     & $6.5\%$                            & $44.9\%$                                 & $5.1\% \pm 0.3\%$  & $37.5\% \pm 2.0\%$ \\
\hline
\textbf{taking photo}   &   $5.9\%$                    & $10.7\%$    & $12.8\%$                           & $22.3\%$                                 & $11.6\% \pm 0.3\%$ & $19.3\% \pm 0.6\%$ \\
\hline
\textbf{walking}        &   $3.4\%$                    & $2.0\%$     & $3.0\%$                            & $28.7\%$                                 & $0.8\% \pm 0.1\%$  & $19.2\% \pm 1.8\%$ \\
\hline
\textbf{using laptop}   &   $27.1\%$                   & $24.2\%$    & $32.9\%$                           & $45.6\%$                                 & $31.5\% \pm 0.6\%$ & $43.0\% \pm 0.6\%$ \\
\hline
\hline
\textbf{mean}           &   $\mathbf{20.5}\%$          & $\mathbf{21.4\%}$ & $\mathbf{27.1\%}$                  & $\mathbf{44.5\%}$                  & $\mathbf{25.7\% \pm 0.2\%}$ & $\mathbf{41.4\% \pm 0.3\%}$ \\
\hline
\end{tabular}
}
\end{center}
\vspace{-2mm}
\caption{Detection performance (inclusion detection average precision criterion) for different methods and levels of supervision.}\label{t:detap_inclusion}
\vspace{-3mm}
\end{table*}

\begin{table*}
\begin{center}
\scalebox{0.78}{
\begin{tabular}{|c|c|c|c|c||c|c|}
\hline
\multirow{2}{*}{\textbf{label}}     & \multicolumn{6}{c|}{\textbf{method/annotations}} \\
\cline{2-7}
%                        &  \textbf{image labels}       & \textbf{eye movements+image labels} & \textbf{labeled bounding boxes} & \textbf{labeled bounding boxes}  \\
                        &  \textbf{CMI-IL-DET}         & \textbf{MI-EYE-DET} & \textbf{CMI-EYE-DET}                & \textbf{BB-DET}                 & \textbf{CMI-EYE-SEQ} & \textbf{BB-SEQ} \\
\hline
\textbf{jumping}        &   $61.9\%$                   &  $64.2\%$           & $69.7\%$                            & $64.5\%$                        & $69.4\% \pm 0.2\%$   & $64.6\% \pm 1.0\%$ \\
\hline
\textbf{phoning}        &   $41.9\%$                   &  $45.2\%$           & $50.4\%$                            & $54.3\%$                        & $49.1\% \pm 0.4\%$   & $56.4\% \pm 0.9\%$ \\
\hline
\textbf{playing guitar} &   $70.2\%$                   &  $69.6\%$           & $80.9\%$                            & $73.4\%$                        & $78.8\% \pm 0.6\%$   & $70.9\% \pm 0.1\%$ \\
\hline
\textbf{reading}        &   $31.3\%$                   &  $30.5\%$           & $40.9\%$                            & $33.6\%$                        & $39.7\% \pm 0.5\%$   & $33.4\% \pm 0.4\%$ \\
\hline
\textbf{riding bike}    &   $74.5\%$                   &  $71.2\%$           & $76.0\%$                            & $84.1\%$                        & $76.0\% \pm 0.1\%$   & $83.1\% \pm 0.5\%$ \\
\hline
\textbf{riding horse}   &   $68.1\%$                   &  $81.7\%$           & $79.4\%$                            & $89.6\%$                        & $84.7\% \pm 0.1\%$   & $87.1\% \pm 0.5\%$ \\
\hline
\textbf{running}        &   $59.0\%$                   &  $63.8\%$           & $60.8\%$                            & $64.7\%$                        & $62.8\% \pm 0.6\%$   & $60.5\% \pm 1.5\%$ \\
\hline
\textbf{taking photo}   &   $27.4\%$                   &  $34.6\%$           & $35.1\%$                            & $30.9\%$                        & $35.2\% \pm 0.4\%$   & $29.9\% \pm 1.0\%$ \\
\hline
\textbf{walking}        &   $26.5\%$                   &  $26.5\%$           & $30.3\%$                            & $47.4\%$                        & $29.2\% \pm 0.5\%$   & $40.5\% \pm 1.4\%$ \\
\hline
\textbf{using laptop}   &   $63.0\%$                   &  $51.2\%$           & $56.3\%$                            & $63.0\%$                        & $55.5\% \pm 0.4\%$   & $62.5\% \pm 0.2\%$ \\
\hline
\hline
\textbf{mean}           &   $\mathbf{52.4}\%$          &  $\mathbf{53.9\%}$  & $\mathbf{58.0\%}$                   & $\mathbf{60.6\%}$               & $\mathbf{58.0\% \pm 0.1\%}$ & $\mathbf{58.9\% \pm 0.4\%}$ \\
\hline
\end{tabular}
}
\end{center}
\vspace{-2mm}
\caption{Action classification performance for different methods and levels of supervision}\label{t:classification}
\vspace{-3mm}
\end{table*}

\subsection{Saccade-and-Fixate Model for Detection}\label{s:sequential_learning_model}

When faced with a recognition task, the human visual system has evolved eye movements to sequentially sample promising image regions through an alternation of saccades and fixations. Inspired by the exceptional, yet unmatched, performance of this biological system, we develop a principled optimal sequential model for image exploration.

At each time step $t$, based on the information gathered so far, our model proposes an image location $\mathbf{z}_t$ that should be sampled. Then, the model updates its internal state $\textbf{s}_t$ based on information contained in the image regions in the neighbourhood of the sampled location. At each time step, the model may terminate the search. In this case, it returns its confidence for the target's presence in the image, together with a region hypothesis for spatial support.

\begin{figure}
\begin{center}
\scalebox{0.28}{
  \includegraphics[viewport=0cm 16.2cm 20.9cm 27.0cm]{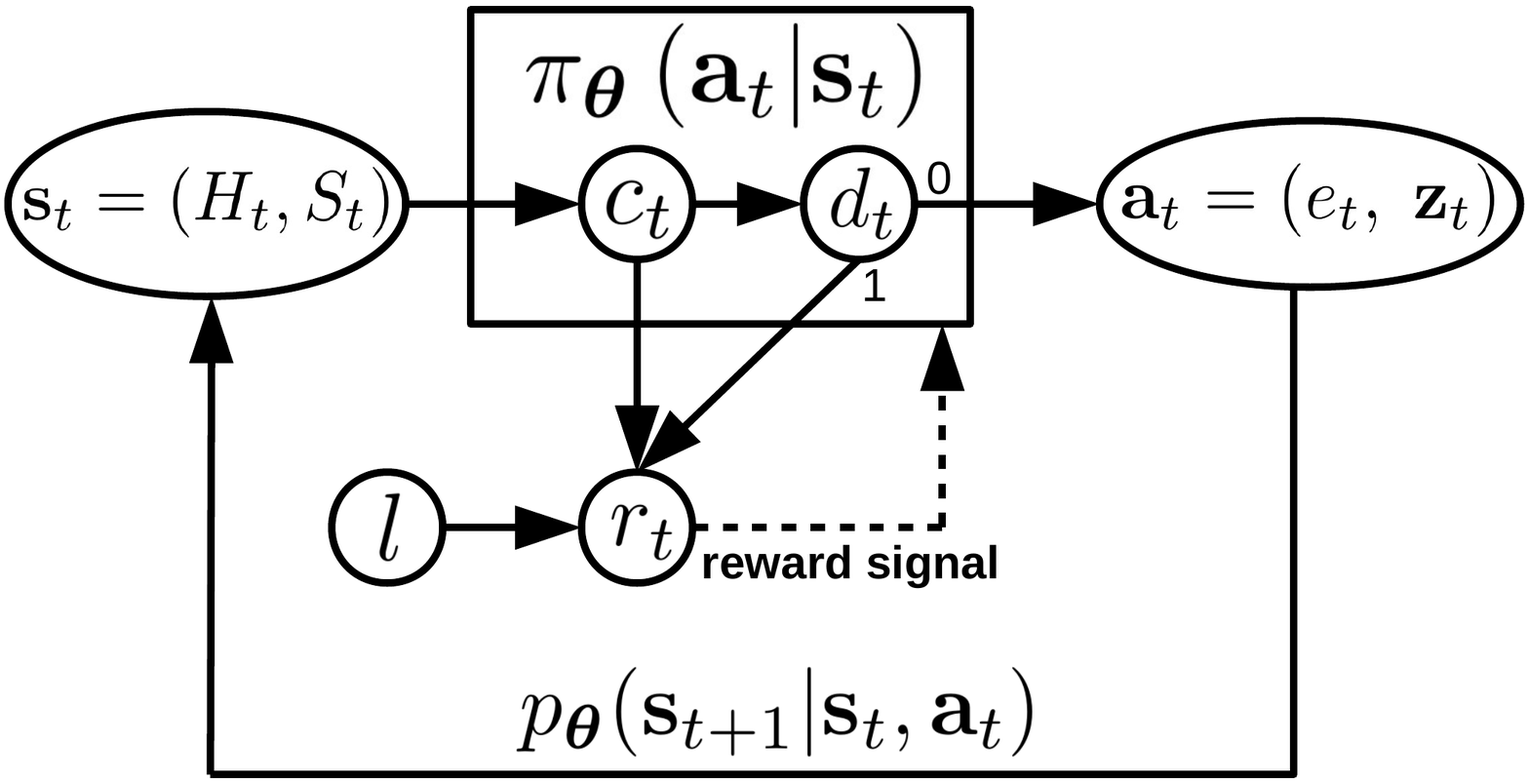}
}
\caption{Sequential detector based on reinforcement learning. At each time step, the model may terminate the search $d_t=1$ depending on the maximal confidence $c_t$ among the history $H_t$ of observed regions (\Algorithm{alg:policy_sampling}), receiving a reward measuring compatibility between confidence $c_t$ and  image label $l$. Otherwise, an evidence region $e_t$ is chosen from the set $H_t\setminus S_t$ of unselected regions and is used to predict the next saccade location $\mathbf{z}_t$. All regions in the neihbourhood of $\mathbf{z}_t$ become observed (\Algorithm{alg:transition}).}\label{fig:seqdet_outline}
\vspace{-5mm}
\end{center}
\end{figure}

We now describe in detail the state and action spaces of our model (\Figure{fig:seqdet_outline}).

\noindent\textbf{Model description:} At each time step $t$, the model keeps track of a set $H_t \subseteq R$ of image regions observed so far, deciding on the appropriate action to take in three steps:

\noindent\underline{\emph{Termination decision:}} The model may decide to terminate search ($d_t=1$). This decision is based on a feature vector of four elements, which consists of the maximum confidence over the set of regions visited so far, the number of saccades already generated, the number of image regions observed so far and a bias:
\begin{equation}
\mathbf{v\left(\mathbf{s}_t\right)}=\left[ \max_{\mathbf{r} \in H_t}{f_{\text{c}}(\mathbf{r})} \;\;\; t \;\;\; \frac{\left\|H_t\right\|}{\left\|R\right\|} \;\;\; 1 \right]^{\top}
\end{equation}

The search termination probability is given by a soft threshold applied to a linear combination of these features, with learned weights $\btheta_{\text{d}}$:
\begin{equation}\label{eq:prob_action}
p_{\btheta}(d_t=1 | \mathbf{s}_t)=\text{sigm}\left[\btheta_{\text{d}}^{\top} \mathbf{v \left(\mathbf{s}_t\right)}\right]
\end{equation}

\noindent where $\text{sigm}(x)=\left(1+e^{-x}\right)^{-1}$ is the sigmoid function.

%\begin{equation}
%f_{\text{a}}(\mathbf{s}_t)=\btheta_{\text{a}}^{\top} \mathbf{v \left(\mathbf{s}_t\right)}
%\end{equation}

\noindent\underline{\emph{Evidence selection:}} If the search is not terminated ($d_t=0$), the model selects an observed image region $e_t \in \left(H_t\setminus S_t\right)$ that provides clues for the location of the target. For example, an image region consisting of the upright torso of a person found in the upper half of the image may offer cues for the presence of a horse in the bottom part.

To capture such effects, we define an evidence function $f_\text{e}:B \rightarrow \mathbb{R}$, $f_{\text{e}}\left(i\right)=\exp\left(\btheta_{\text{e}}^{\top} \mathbf{g}(\mathbf{r}_i)\right)$ that evaluates the informativeness of image region $i$ with respect to the target location. We select the region $e_t$ from a multinomial distribution defined by the evidence function over the set $H_t \setminus S_t$ of image regions not selected during previous steps:
\begin{equation}\label{eq:prob_evidence}
p_{\btheta}(e_t  | \mathbf{s}_t)=\frac{f_\text{e}(e_t)}{\sum_{i \in H_t \setminus S_t} f_\text{e}(i)}
\end{equation}

\noindent\underline{\emph{Location selection:}} Once selected, the evidence region $e_t$ is used to define a 2d Gaussian probability distribution for the next fixation location $\mathbf{z}_t \in \mathbb{R}^2$:

\begin{equation}\label{eq:prob_position}
p_{\btheta}(\mathbf{z}_t | \mathbf{s}_t, e_t)=N\left(\cdot,f_{\text{p}}(e_t),\btheta_{\Sigma}\right)
\end{equation}

\noindent where the center $f_{\text{p}}(e_t)$ is based on a linear combination of the evidence region features $g(e_t)$:
\vspace{-1mm}
\begin{equation}
f_{\text{p}}(i)=\left[\btheta_{\text{p}}^{\top} g(\mathbf{r}_i)\right] \cdot \frac{\mathbf{x}_2(\mathbf{r}_i)-\mathbf{x}_1(\mathbf{r}_i)}{2}+\frac{\mathbf{x}_1(\mathbf{r}_i)+\mathbf{x}_2(\mathbf{r}_i)}{2}
\end{equation}

\vspace{-1mm}

We make the position function invariant to the scale of the image region $\mathbf{r}_i$ by normalizing with respect to its bounding box, defined by the top-left and bottom right corners $\mathbf{x}_2(\mathbf{r}_i)$ and $\mathbf{x}_1(\mathbf{r}_i)$.

%Once selected, the evidence region $e_t$ defines a probability distribution for the next fixation location. This distribution is a 2D Gaussian with standard deviation $\Sigma$ and center define by evaluating a position function $f_{\text{p}}:R \rightarrow \mathbf{\mathbb{R}}^2$.

The policy of our agent can be expressed as:
\vspace{-1mm}
\begin{equation}
\pi_{\btheta}\left(\mathbf{a}_t | \mathbf{s}_t\right)=p_{\btheta}\left(d_t=0|\mathbf{s}_t\right) p_{\btheta}\left(e_t | \mathbf{s}_t\right) p_{\btheta}\left(\mathbf{z}_t | \mathbf{s}_t, e_t\right)
\end{equation}

\vspace{-1mm}

The model is executed by repeated sampling of the policy $\pi_{\btheta}\left(\mathbf{a}_t | \mathbf{s}_t\right)$, until termination ($d_t=1$) (\Algorithm{alg:policy_sampling}). At each step the state $\mathbf{s}_t$ is updated according to the action $\mathbf{a}_t$ (\Algorithm{alg:transition}). When the search is finished, the observed region $\mathbf{r}_k$ with maximum confidence $c_t$ is returned as detector output.

\noindent\textbf{Training algorithm:} In our weakly supervised setting, we are given a set of images, represented as bags of regions $B_j$ with associated global image labels $l_j$, together with confidence function $f_c$ trained to be maximal at target locations, either with strong or weak supervision, \cf\Section{s:confidence_learning_method}. We wish to find the model parameters $\btheta$ maximizing image-level classification accuracy based on the confidence $c_t$ at the last step (when $d_t=1$), while at the same time  minimize the number of segment evaluations. Note that because the confidence response $c_t$ of the agent is the maximal confidence function response, we also encourage good detection performance.

We formulate our training objective in a reinforcement learning framework. Our reward function is sensitive to the confidence at the final state and incurs a penalty for each saccade the agent makes:
\vspace{-1mm}
\begin{equation}
r_t=
\begin{cases}
-\alpha & \text{if}\; d_t=0\\
\min(c_t,1) & \text{if}\; d_t=1 \wedge l=1\\
\max(c_t,-1) & \text{if}\; d_t=1 \wedge l=-1
\end{cases}
\end{equation}

\vspace{-1mm}

\noindent where $l$ is the label associated with the training image over which the model was executed and $\alpha$ is a penalty paid by the model for each saccade.

During training, we maximize the expected reward function on the training set, defined as:
\vspace{-1mm}
\begin{equation}\label{eq:objective}
F(\btheta)=\mathbb{E}_{p_{\btheta}(\mathbf{s})} \left[\sum_{t=1}^{\left|\mathbf{s}\right|} r_t\right] + \frac{\lambda}{2} \btheta^{\top}\btheta
\end{equation}

\vspace{-1mm}

\noindent where $\mathbf{s}$ represents a variable length sequence of states, sampled by running the model (\Algorithm{alg:policy_sampling} and \ref{alg:transition}), starting from an initial state $\mathbf{s}_0=(H_0,S_0)$ and $\lambda$ is an L2 regularizer. We set $H_0$ to the set of segments observed by fixating the image center and $S_0$ to $\emptyset$.

The gradient of the expected reward can be approximated as\cite{Williams-ml-1992,SuttonBarto-1998}:
\begin{equation}
\nabla_{\btheta} F(\btheta)=\frac{1}{M}{\sum_{i=1}^{M} \sum_{t=1}^{\left|\mathbf{s}^{i}\right|}} \nabla_{\btheta} \text{log} \pi_{\btheta}(\mathbf{a}^{i}_t | \mathbf{s}^{i}_t) \left[\sum_{t=1}^{\left|\mathbf{s}\right|} r_t\right] + \lambda\btheta
\end{equation}

\noindent where $\mathbf{s}^{i}$, $\mathbf{a}^{i}_t$ and $r^{i}_t$, $i=1\dots M$, represent sequences of states, actions and corresponding rewards, sampled by model simulation.

\section{Experimental Results and Discussion}

In this section, we first discuss our experimental methodology and results for learning confidence functions (\Section{s:confidence_learning}) and optimal search strategies for these functions (\Section{s:sequential_learning}).

\subsection{Learning Confidence Functions}\label{s:confidence_learning_results}

\begin{figure*}
\begin{center}
\scalebox{0.90}{
\scalebox{0.169}{
	\includegraphics[viewport=0cm 0cm 9.4cm 14.1cm]{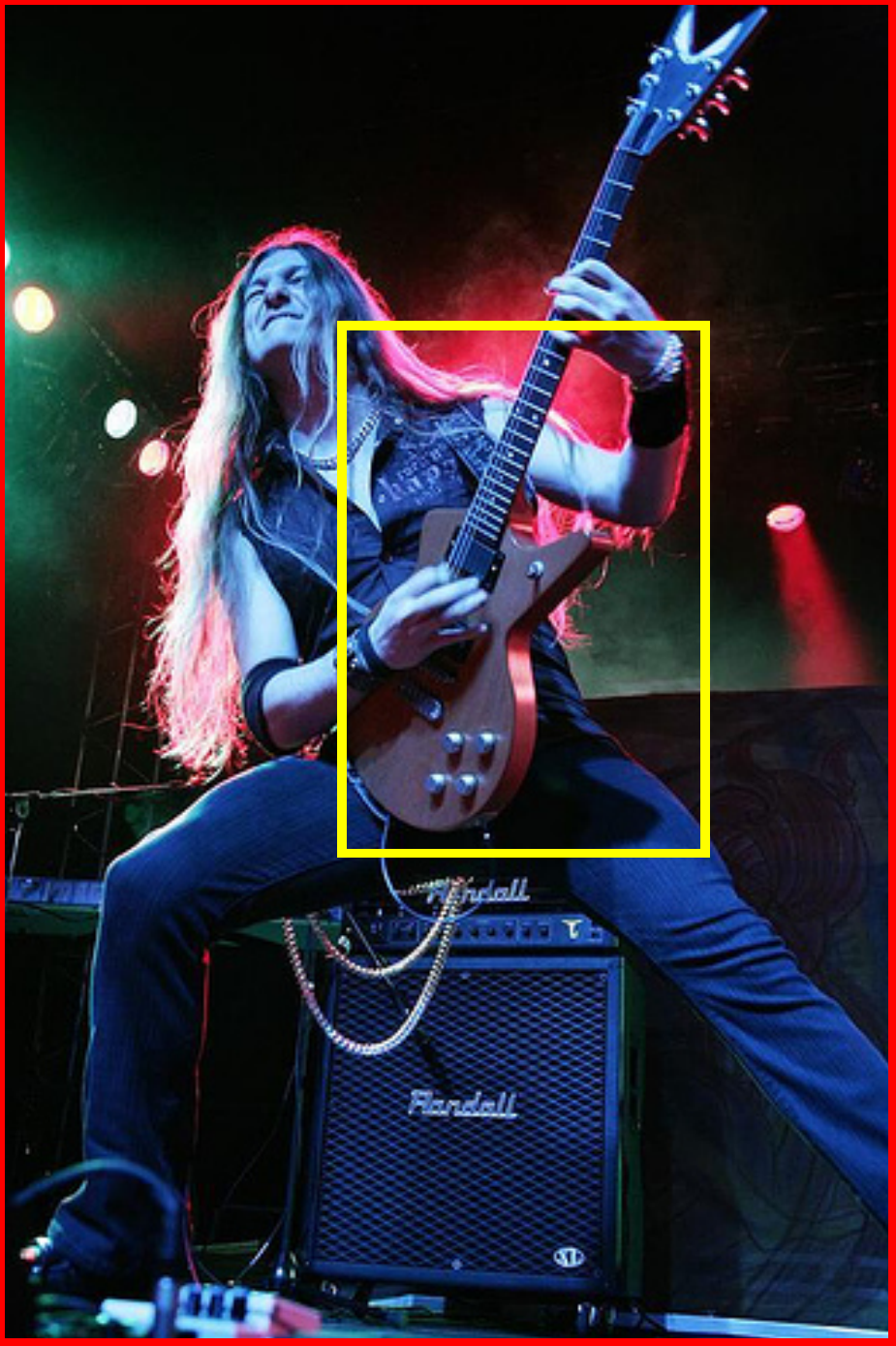}
}%
\hspace{0.3mm}%
\scalebox{0.169}{
	\includegraphics[viewport=0cm 0cm 9.4cm 14.1cm]{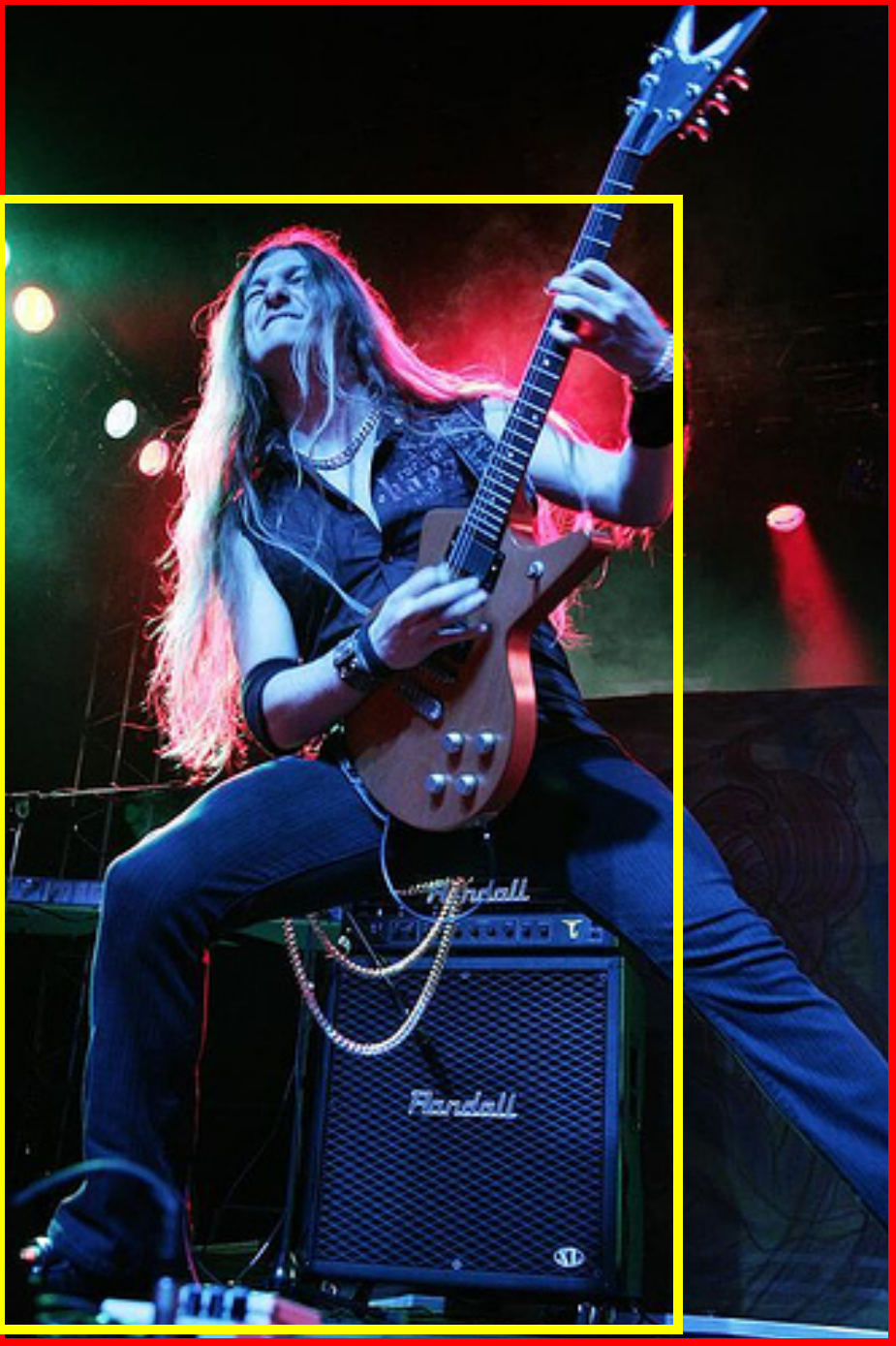}
}%
\hspace{1.0mm}%
\scalebox{0.169}{
	\includegraphics[viewport=0cm 0cm 11.4cm 14.1cm]{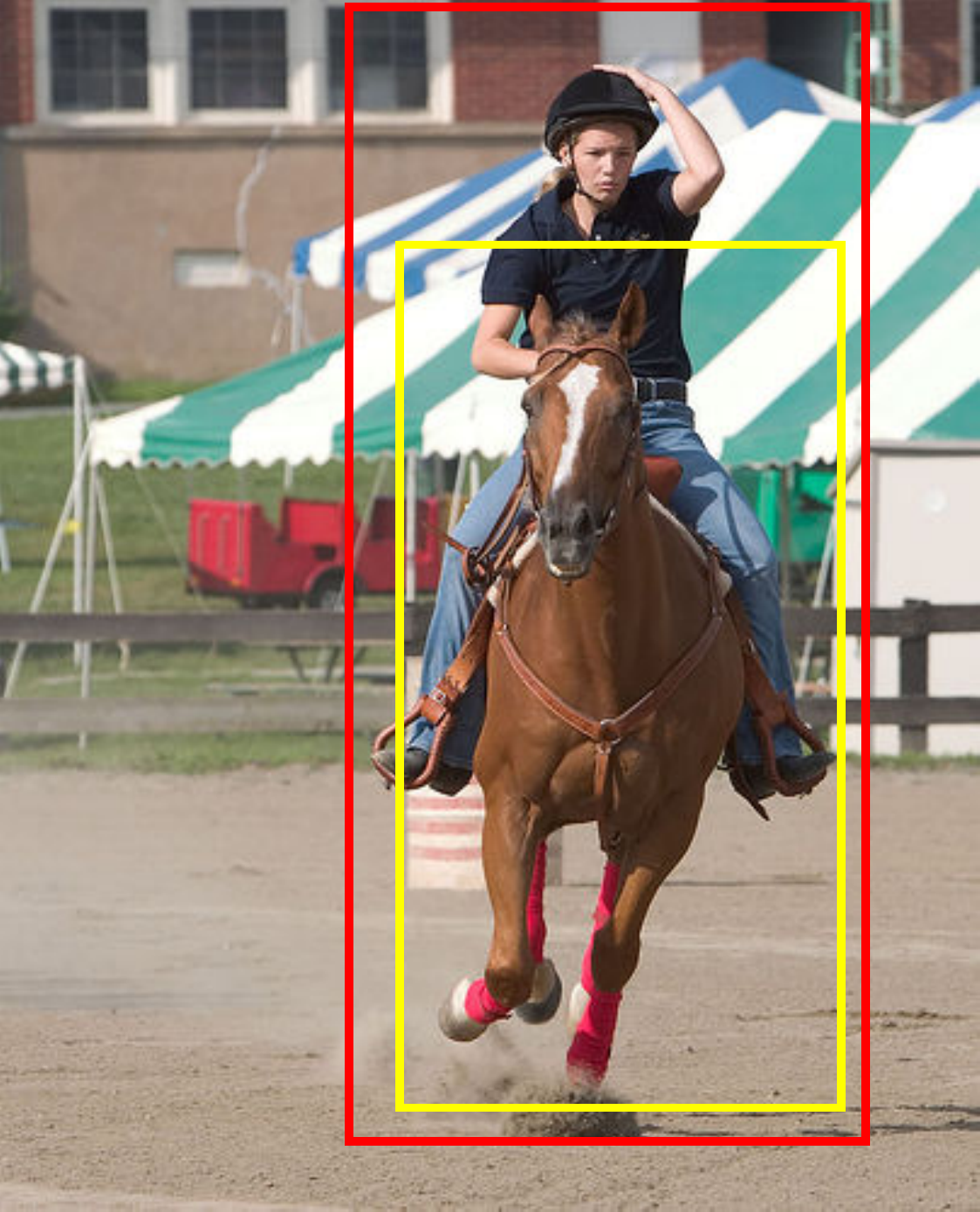}
}%
\hspace{0.3mm}%
\scalebox{0.169}{
	\includegraphics[viewport=0cm 0cm 11.4cm 14.1cm]{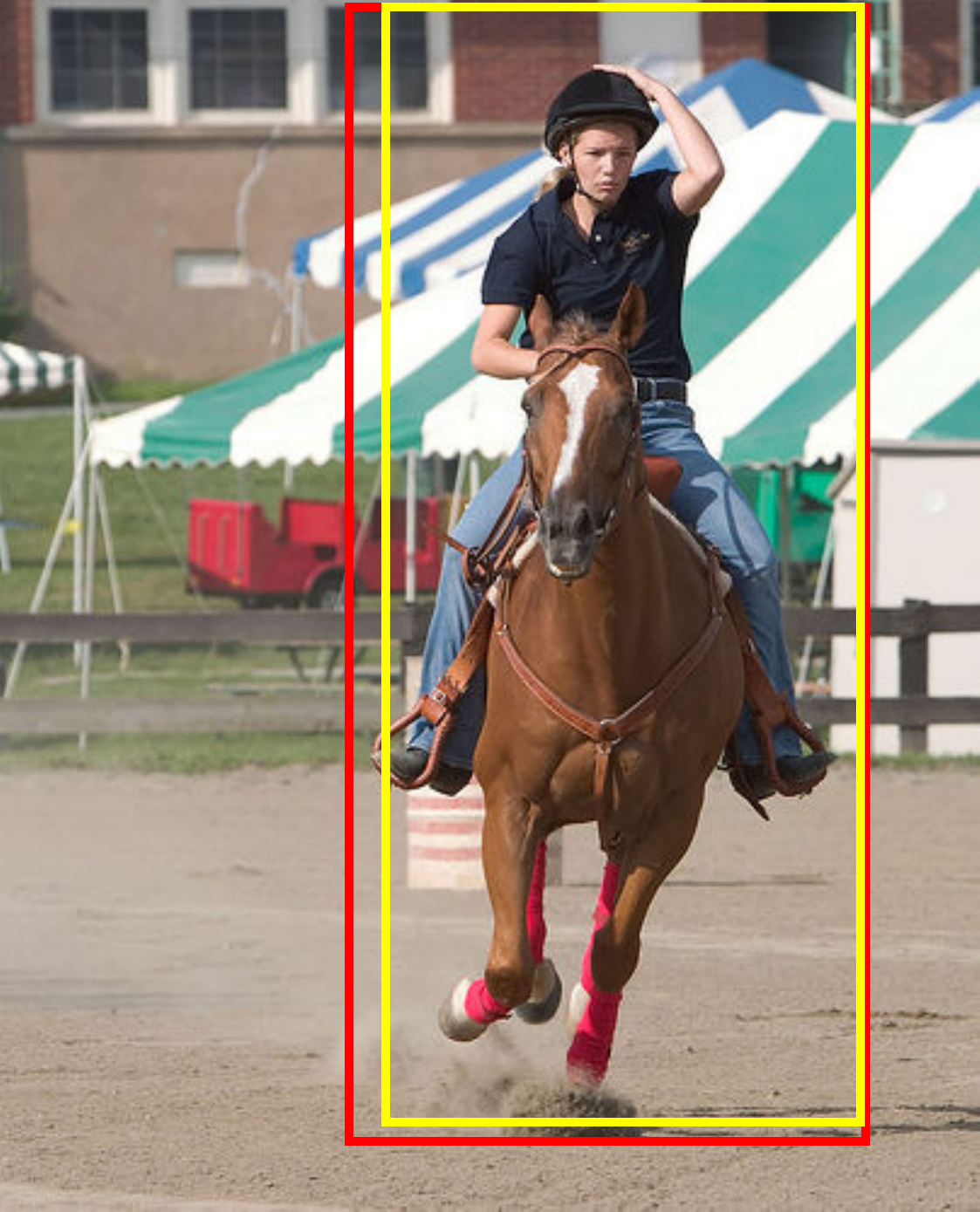}
}%
\hspace{1.0mm}%
\scalebox{0.169}{\includegraphics[viewport=0cm 0cm 10.6cm 14.1cm]{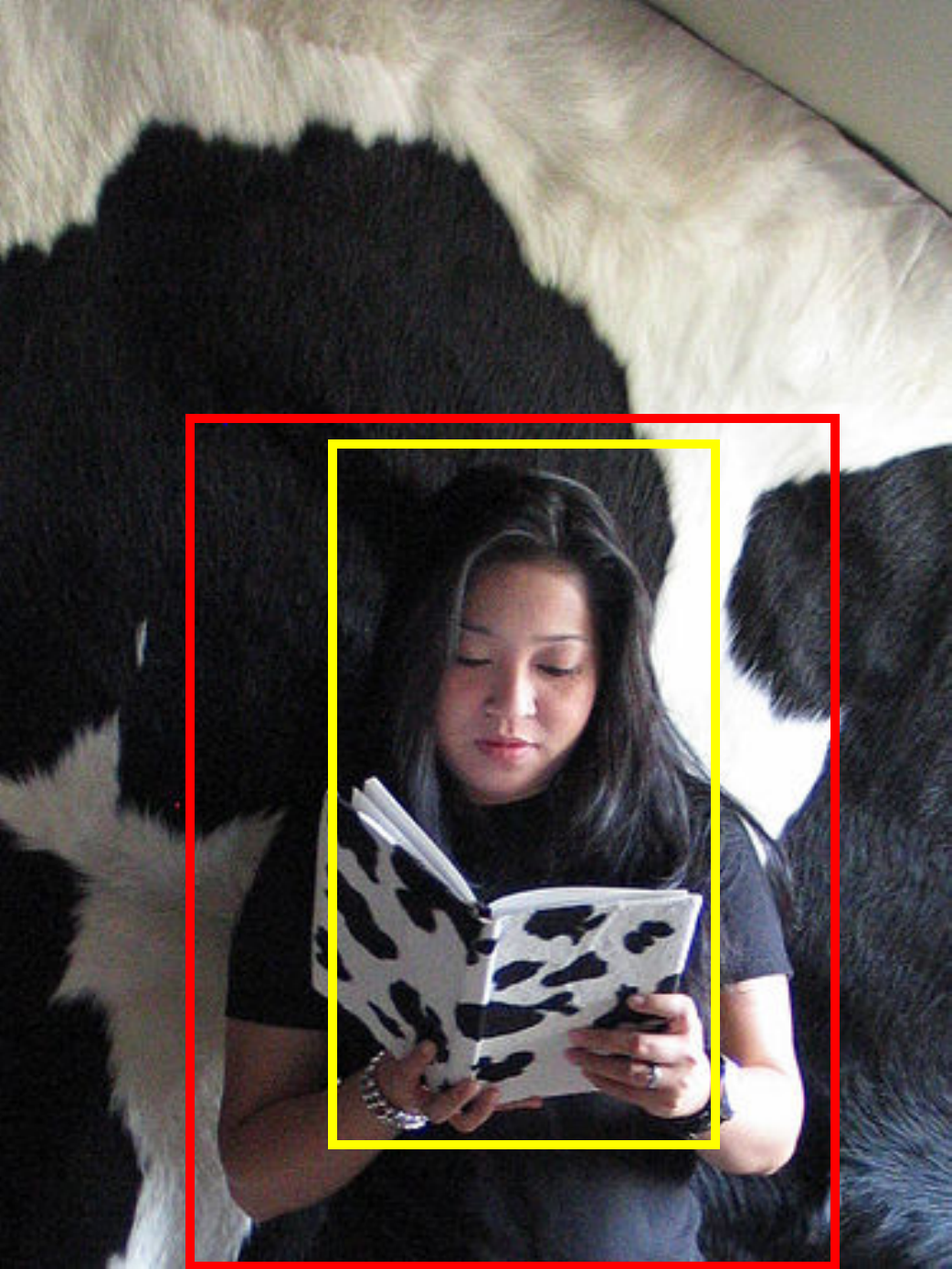}}%
\hspace{0.3mm}%
\scalebox{0.169}{\includegraphics[viewport=0cm 0cm 10.6cm 14.1cm]{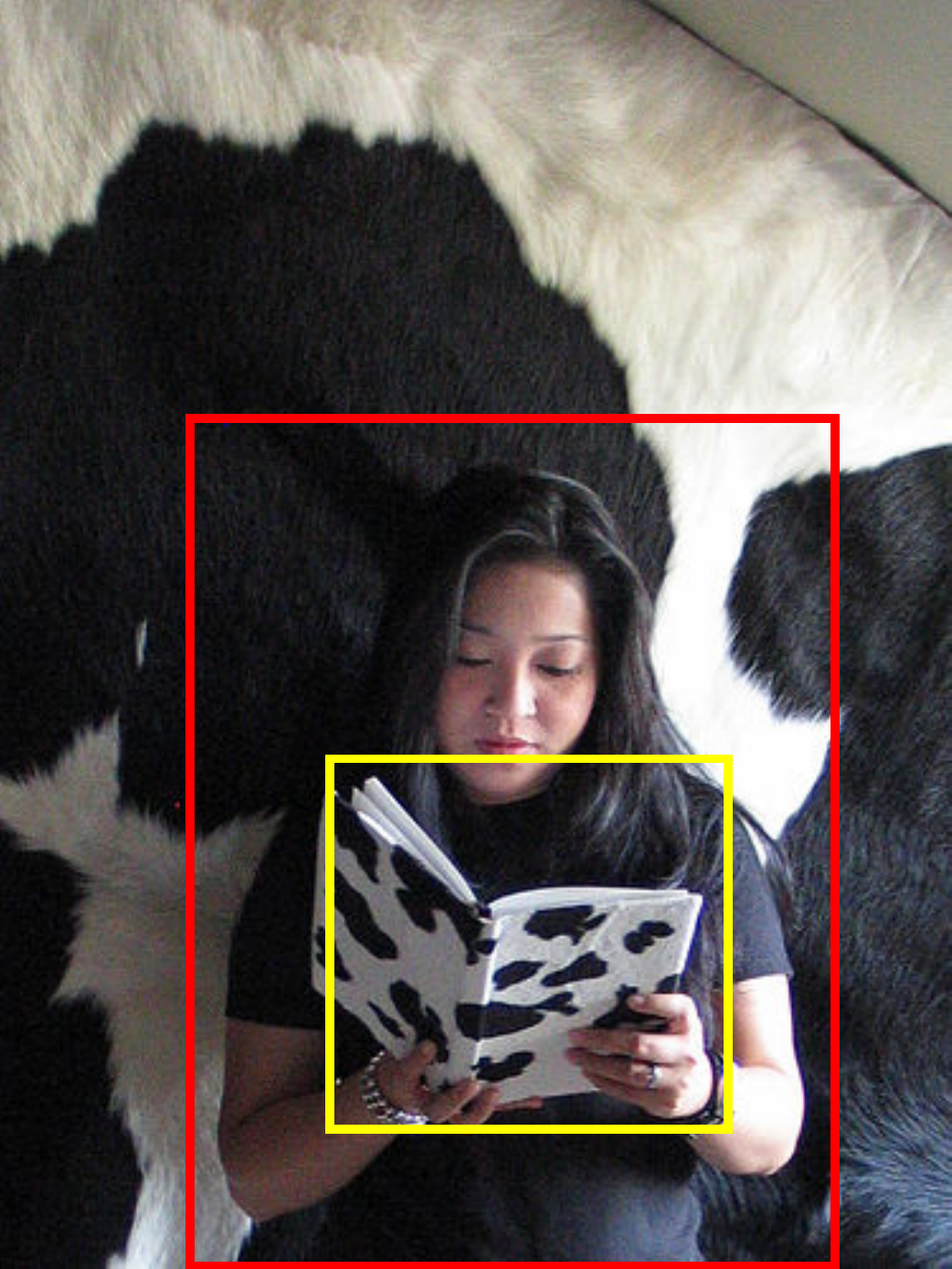}}%
\hspace{1.0mm}%
\scalebox{0.2235}{\includegraphics[viewport=0cm 0cm 14.1cm 10.6cm]{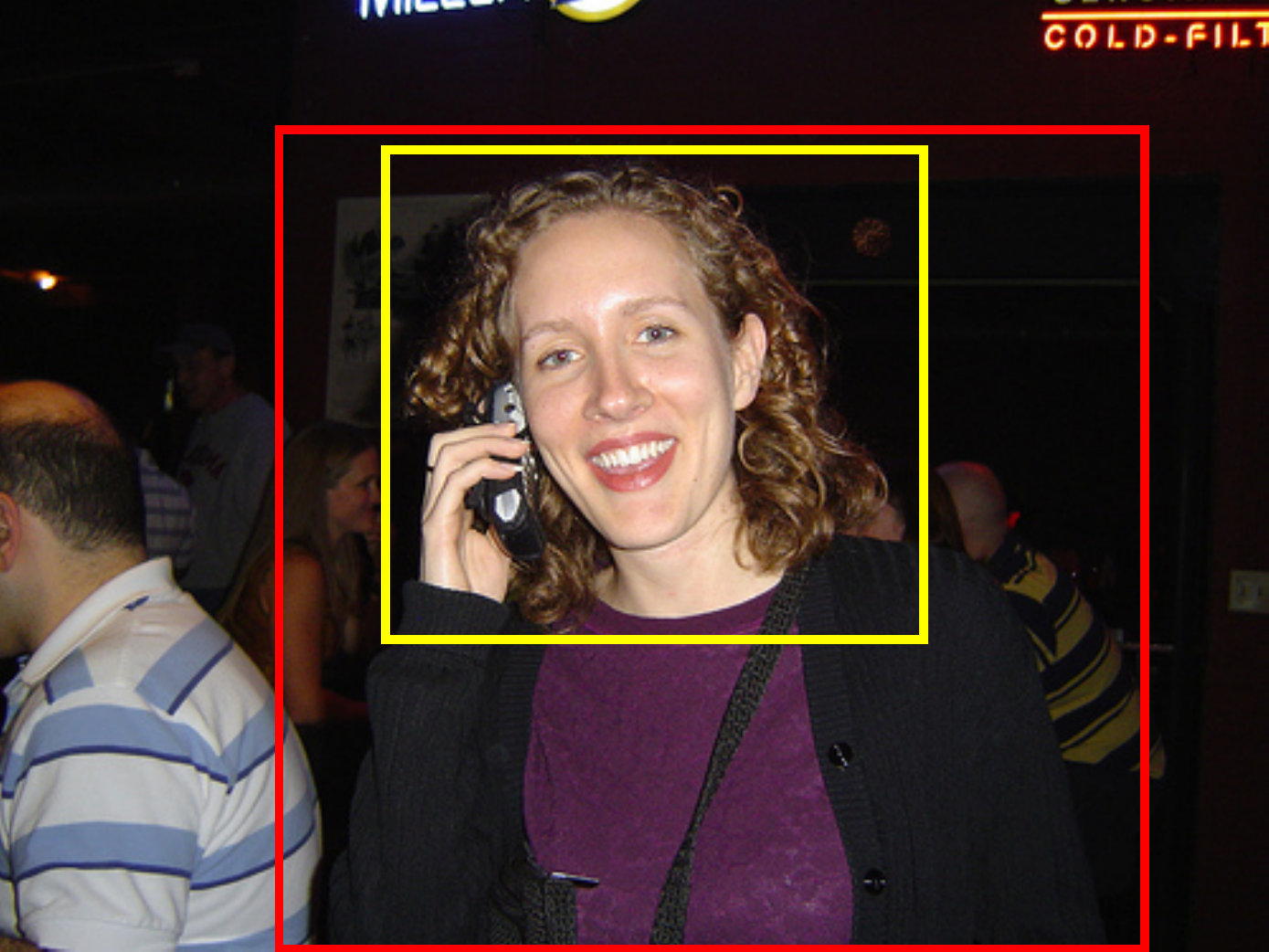}}%
\hspace{0.3mm}%
\scalebox{0.2235}{\includegraphics[viewport=0cm 0cm 14.1cm 10.6cm]{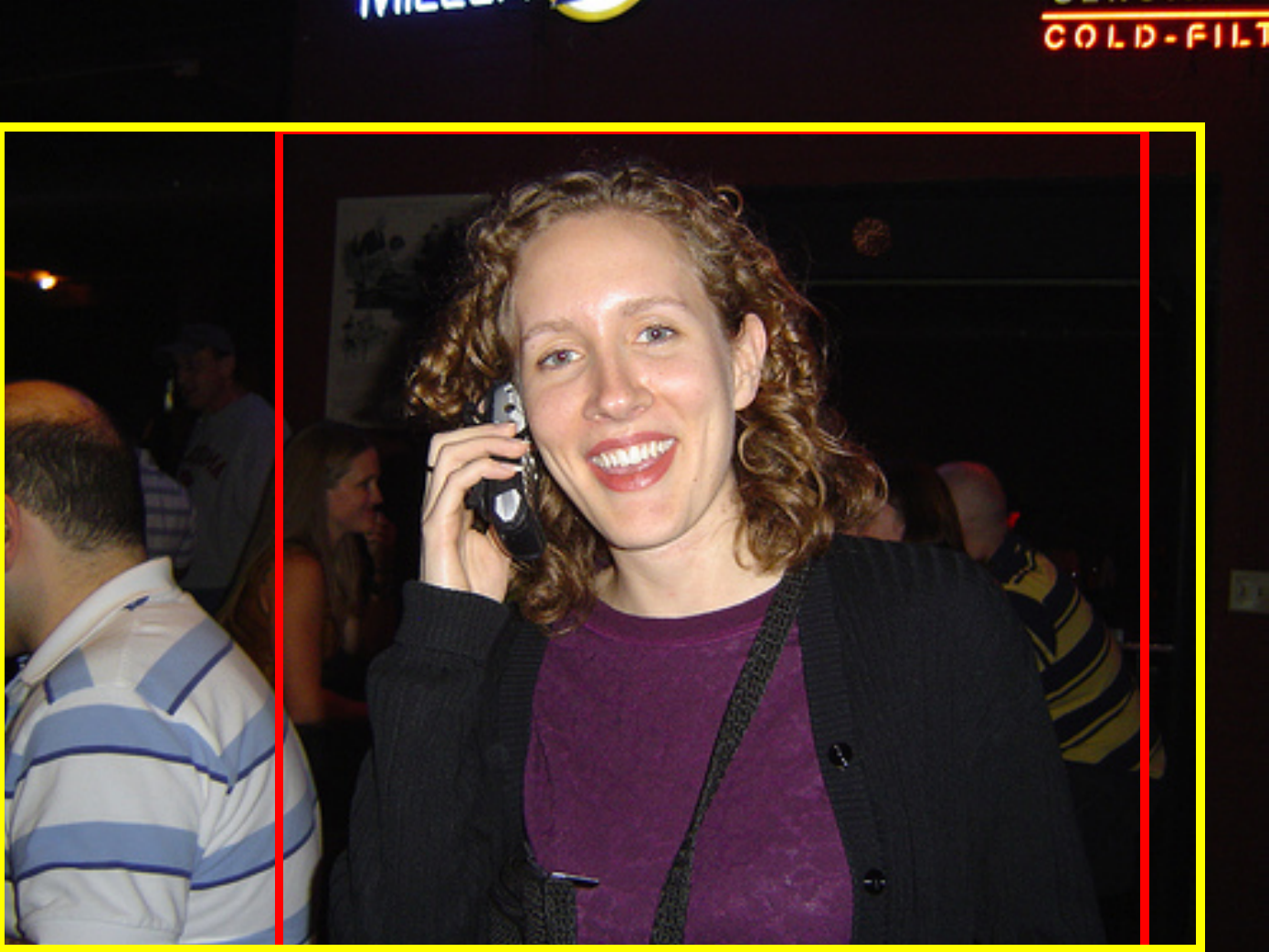}}%
}%
\caption{Detection responses (yellow) vs. ground truth (red), for the CMI-DET-EYE pipeline (left in pair) and the DET-BB pipeline (right in pair).
Note that classifiers often tend to respond to a sub-region of the ground bounding box which defines the action-specific pattern.
This trend is more pronounced for weakly supervised pipelines (\textit{e.g.} CMI-DET-EYE) than fully supervised ones (DET-BB),
as also measured numerically by our inclusion metric (see \tableref{t:detap_inclusion}). Please see supplementary material for more results.}\label{fig:predbb}
\vspace{-5mm}
\end{center}
\end{figure*}

\begin{table*}
\begin{center}
\scalebox{0.78}{
\begin{tabular}{|c|c|c|c||c|c|c|}
\hline
\multirow{3}{*}{\textbf{label}} & \multicolumn{3}{c||}{\textbf{SEQ-DET-EYE}}  & \multicolumn{3}{c|}{\textbf{SEQ-DET-BB}} \\
\cline{2-7}
                        &  \multirow{2}{1.5cm}{\centering\textbf{running time (s)}}  & \multirow{2}{1.8cm}{\centering\textbf{evaluated segments}} & \multirow{2}{1.2cm}{\centering\textbf{speedup}} & \multirow{2}{1.6cm}{\centering\textbf{running time (s)}} & \multirow{2}{1.8cm}{\centering\textbf{evaluated segments}} & \multirow{2}{1.2cm}{\textbf{speedup}} \\
                        &  & & & & & \\
\hline
\textbf{jumping}        &  $19.8 \pm 0.7$              &  $44.4 \pm 0.3\%$   & $2.1$ & $27.1 \pm 0.7$       & $51.7\% \pm 0.2\%$ & $1.6$ \\
\hline
\textbf{phoning}        &  $19.6 \pm 0.8$              &  $44.1 \pm 0.4\%$   & $2.2$ & $27.2 \pm 0.7$       & $52.3\% \pm 0.2\%$ & $1.6$ \\
\hline
\textbf{playing guitar} &  $19.5 \pm 0.6$              &  $43.9 \pm 0.3\%$   & $2.2$ & $22.5 \pm 0.2$       & $38.1\% \pm 0.0\%$ & $1.9$ \\
\hline
\textbf{reading}        &  $19.7 \pm 0.8$              &  $44.3 \pm 0.3\%$   & $2.2$ & $26.6 \pm 0.8$       & $50.4\% \pm 0.2\%$ & $1.6$ \\
\hline
\textbf{riding bike}    &  $18.4 \pm 0.2$              &  $38.1 \pm 0.1\%$   & $2.3$ & $28.2 \pm 0.7$       & $55.6\% \pm 0.2\%$ & $1.5$ \\
\hline
\textbf{riding horse}   &  $19.7 \pm 0.9$              &  $44.1 \pm 0.3\%$   & $2.2$ & $27.5 \pm 0.7$       & $53.1\% \pm 0.2\%$ & $1.6$ \\
\hline
\textbf{running}        &  $19.8 \pm 0.8$              &  $44.5 \pm 0.3\%$   & $2.2$ & $30.2 \pm 1.2$       & $61.0\% \pm 0.4\%$ & $1.4$ \\
\hline
\textbf{taking photo}   &  $19.5 \pm 0.6$              &  $41.4 \pm 0.3\%$   & $2.2$ & $28.0 \pm 0.8$       & $54.6\% \pm 0.2\%$ & $1.5$ \\
\hline
\textbf{walking}        &  $19.1 \pm 0.6$              &  $42.9 \pm 0.2\%$   & $2.3$ & $30.9 \pm 0.2$       & $63.4\% \pm 0.3\%$ & $1.4$ \\
\hline
\textbf{using laptop}   &  $20.0 \pm 0.9$              &  $45.6 \pm 0.4\%$   & $2.1$ & $24.8 \pm 0.3$       & $45.0\% \pm 0.4\%$ & $1.7$ \\
\hline
\hline
\textbf{mean}           & $\mathbf{19.5 \pm 0.3}$      &  $\mathbf{43.3 \pm 0.2}$ & $\mathbf{2.2}$ & $\mathbf{27.3} \pm 0.5$ & $\mathbf{52.5\% \pm 0.3\%}$ & $\mathbf{1.6}$ \\
\hline
\end{tabular}
}
\end{center}
\caption{Computational cost comparison for learned optimal search strategies (CMI-SEQ-EYE and CMI-SEQ-BB). Speedups are measured with respect to exhaustive search (CMI-DET-IL, CMI-DET-BB, CMI-DET-EYE), which takes, on average, $43.0 \pm 0.1$ seconds per image.}\label{t:seq_det_speedup}
\end{table*}

Our confidence function (\textbf{CMI-EYE-DET}) is trained on the set of image regions fixated by human subjects, as described in \Section{s:fixation_studies}. For positive images, where bags consist of fixated segments, we create additional negative bags consisting of all segments not overlapping any fixated segment. We use threholds $T_S=T_F=0.2$ to define the fringe and subregions \cf \eqref{eq:subseg} and \eqref{eq:fringe}. For initialization, we generate 30 random label assignments $\mathbf{y}$, with the ratio of positive/negative labels uniformly sampled from the range $\left[0.1,1.0\right]$. Our model generally converges in less than 20 iterations. We evaluate the SVM-C parameter on the validation set, and re-train on trainval. We then evaluate our confidence function on all segments extracted from each test image. Detector responses for each image are generated by non-max suppression at overlap threshold $0.2$.

\noindent\textbf{Features:} We use the deep convolutional neural network of Krizhevsky\textit{et al.}\cite{KrizhevskyEtAl-nips-2012} as feature extractor $g$. Given an image segment, we first compute its bounding box and enlarge it by $10\%$. We then apply the neural network twice to produce our feature descriptor. First, we apply it on the contents of the bounding box (after proper scaling). Second, to capture context, we consider the entire image, mask the bounding box by filling it with its mean color, rescaled and passed through the network. For each call we record the output of the last fully connected layer, and end up with two vectors of 4096 real values. Additionally, we build a representation of the bounding box that consists of its normalized image coordinates, size and aspect ratio. Our final feature vector consists of the two concatenated neural response vectors and the bounding box descriptor, and has 8204 entries.

\noindent\textbf{Baselines:} We compare our method to three baselines:

\noindent\underline{\emph{Supervised learning (\textbf{BB-DET}):}} We train a confidence function where positive examples are the image segments with the highest target overlap (see \tableref{t:segment_filtering}, column 3), the fringe of these segments is removed from the training set, and the rest of the segments are negative examples.

\noindent\underline{\emph{Multiple instance learning (\textbf{MI-EYE-DET}):}} To capture the contribution of our topological constraints to model performance, we train our confidence function on the set of fixated segments in a similar fashion to CMI-EYE-DET, without constraints (4.2), (4.3), restricting the range of possible image labels in constraint (4.1) to $\left\{-1,1\right\}$.

\noindent\underline{\emph{Image labels alone (\textbf{CMI-IL-DET}):}} To estimate the power of eye movements as a supervisory signal, we also investigate the setup in which only image labels are available at training time. To this end, we train our model on bags composed of the entire segment pool for the corresponding image, as returned by CPMC.

\noindent\textbf{Evaluation:} We first evaluate our models under the VOC detection average precision metric, in which a segment is considered a hit if its bounding box overlaps the ground truth bounding box with at least $0.5$ overlap.

Note however that our ground truth bounding boxes contain both the person and the manipulated object. The distinctive pattern that defines an action is often a subregion within this bounding box, \textit{e.g.} the person's hands grasping the guitar consistent with playing or the person's head and the mobile phone close to her ear (see \Figure{fig:predbb}). While precisely defining annotations for these patterns is not trivial, we also evaluate our methods on a metric that counts detector responses falling within the predicted bounding box as correct, regardless of the overlap. Under this metric, a segment is a hit if its bounding box falls within the subregion set \eqref{eq:subseg} of the ground truth bounding box, with $T_S=0.5$.

We also evaluate image level classification, by computing the maximum confidence of the trained detector for each image, and reporting classification average precision over the images in the test set.

\noindent\textbf{Results:} Our method (CMI-EYE-DET) outperforms both baselines under all metrics (tables \ref{t:detap}, \ref{t:detap_inclusion} and \ref{t:classification}). There is a larger performance gap between supervised and weakly supervised approaches when measured under the average precision metric, than under a less restrictive inclusion metric. This suggests, that weakly supervised approaches recover invariant patterns within the bounding box, although, at least in the absence of prior knowledge, cannot easily estimate the full extent of the actor and person. Our visualizations of the results (\Figure{fig:predbb}) also support this intuition. 

The contribution of topological constraints to the multiple instance learning quality is substantial under all metrics. Additionally, leveraging human eye movement annotations produces image classifiers nearly as good as those learned under full supervision. The standard MI-EYE-DET pipeline fails to exploit the human eye movement training signal for image classification, showing that our proposed topological constraints are crucial for this task.

\subsection{Learning Optimal Search Strategies}\label{s:sequential_learning_results}

We train sequential detection search models for the confidence function learned using eye movements (CMI-EYE-SEQ) and bounding boxes (BB-SEQ). Note that for both cases, our search model is learned using image labels alone. We optimize our objective function \eqref{eq:objective} using a BFGS optimizer and we set the $\lambda$ regularizer to maximize the expected reward on the validation set. We use 8 random initializations of the model parameters $\btheta$ and run the model until convergence, which generally takes less than 30 iterations.

\noindent\textbf{Evaluation:} Our detector is run on the test set using the metrics in \Section{s:confidence_learning}. We also measure the number of evaluated segments and the total computational time, using a C++/Matlab implementation on an Intel Xeon E5 2660 2.20GHz CPU.

\noindent\textbf{Results:} We find that our learned optimal search model only requires feature extraction for approximately half of the segments (\tableref{t:seq_det_speedup}), while leaving image level classification performance (\tableref{t:classification}) at nearly the same levels as in the corresponding pipelines based on exhaustive evaluation (CMI-EYE-DET and BB-DET). At the same time, detection performance is only slightly affected compared to exhaustive search. The total computational time is also improved. Note that our numbers account for both CPMC segment extraction and the running time of the search model itself, and are overall significantly improved. In all experiments we use an optimized version of CPMC, which operates on a reduced search space formed by superpixels, 
%instead of individual pixels (
offering significant speedups with respect to standard CPMC.

\section{Conclusions}

We have presented novel, general, weakly supervised segmentation-based methods to learn accurate and efficient detection models in static images. In contrast to methods trained using ground truth bounding box or segment annotations, our detection response model leverages novel multiple instance learning techniques with topological constraints in order to learn accurate confidence functions and image classifiers using eye movement data. Additionally, we develop novel sequential models, in order to achieve optimal, efficient search strategies for detection based on reinforcement learning. In extensive experiments, we show that our proposed methodology achieves significant progress in terms of accuracy and speed, under weak supervision.
%Our results underline the potential of human eye movements both as a training signal and as a source of insights for designing computer vision systems.

%$p_{\btheta}(\mathbf{s}_{t+1}|\mathbf{s}_{t},\mathbf{a}_{t})$

%\clearpage
%\newpage

{\small
\bibliographystyle{ieee}
\bibliography{arxiv_detection}
}

\end{document}